\documentclass[11pt]{article}
\usepackage{arxiv}
\usepackage{url}
\usepackage{latexsym}
\usepackage{booktabs}
\usepackage{multirow}
\usepackage{graphicx}
\usepackage{todonotes}
\usepackage{amssymb}
\usepackage{amsmath}
\usepackage{array}
\usepackage{hyperref}
\usepackage{makecell}
\usepackage{rotating}
\usepackage{longtable}
\usepackage{authblk}
\usepackage{doi}

\pdfoutput=1

\usepackage{times}
\usepackage{latexsym}

\usepackage[T1]{fontenc}

\usepackage[utf8]{inputenc}

\usepackage{microtype}

\title{medBERT.de: A Comprehensive German BERT Model for the Medical Domain}

\author[1,2,3,9,*]{Keno K. Bressem}
\author[4,*]{Jens-Michalis Papaioannou}
\author[4,*]{Paul Grundmann}
\author[5]{Florian Borchert}
\author[6,7]{Lisa C. Adams}
\author[4]{Leonhard Liu}
\author[2]{Felix Busch}
\author[2]{Lina Xu}
\author[2]{Jan P. Loyen}
\author[2]{Stefan M. Niehues}
\author[8]{Moritz Augustin}
\author[8]{Lennart Grosser}
\author[7]{Marcus R. Makowski}
\author[1,9,10]{Hugo JWL. Aerts}
\author[4]{Alexander Löser}

\affil[1]{Artificial Intelligence in Medicine (AIM) Program, Mass General Brigham, Harvard Medical School, Boston, MA, USA,  \authorcr Email: \tt \{kbressem, HAerts\}@bwh.harvard.edu \vspace{0.3cm}} 
\affil[2]{Charité – Universitätsmedizin Berlin, corporate member of Freie Universität Berlin and Humboldt-Universität zu Berlin, Institute for Radiology, Berlin, Germany  \vspace{0.3cm}}
\affil[3]{Berlin Institute of Health at Charité – Universitätsmedizin Berlin, Germany \vspace{0.3cm}}
\affil[4]{Berliner Hochschule für Technik (BHT), Berlin, Germany \authorcr Email:  \tt \{michalis.papaioannou, pgrundmann, leonhard.liu, aloeser\}@bht-berlin.de \vspace{0.3cm}}
\affil[5]{Digital Health Center, Hasso Plattner Institute, University of Potsdam, Germany, \authorcr Email: \tt \{florian.borchert\}@hpi.de \vspace{0.3cm}}
\affil[6]{Department of Radiology, Stanford University School of Medicine, Stanford, CA, USA \authorcr Email: \tt \{lcadams\}@stanford.edu \vspace{0.3cm}}
\affil[7]{Department of Diagnostic and Interventional Radiology, School of Medicine and Klinikum Rechts der Isar, Technical University of Munich, Munich, Germany, \authorcr Email: \tt \{marcus.makowski\}@tum.de  \vspace{0.3cm}}
\affil[8]{Tiplu GmbH, Hamburg, Germany \authorcr Email: \tt \{m.augustin, l.grosser\}@tiplu.de \vspace{0.6cm}}
\affil[9]{Departments of Radiation Oncology and Radiology, Dana-Farber Cancer Institute and Brigham and Women’s Hospital, Boston, MA, USA \vspace{0.3cm}}
\affil[10]{Radiology and Nuclear Medicine, CARIM \& GROW, Maastricht University, Maastricht, the Netherlands \vspace{0.3cm}}
\affil[*]{Contributed equally}

\begin{document}

\maketitle

\begin{abstract}
This paper presents medBERT.de, a pre-trained German BERT model specifically designed for the German medical domain. The model has been trained on a large corpus of 4.7 Million German medical documents and has been shown to achieve new state-of-the-art performance on eight different medical benchmarks covering a wide range of disciplines and medical document types. In addition to evaluating the overall performance of the model, this paper also conducts a more in-depth analysis of its capabilities. We investigate the impact of data deduplication on the model's performance, as well as the potential benefits of using more efficient tokenization methods. Our results indicate that domain-specific models such as medBERT.de are particularly useful for longer texts, and that deduplication of training data does not necessarily lead to improved performance. Furthermore, we found that efficient tokenization plays only a minor role in improving model performance, and attribute most of the improved performance to the large amount of training data. To encourage further research, the pre-trained model weights and new benchmarks based on radiological data are made publicly available for use by the scientific community.

\end{abstract}

\section{Introduction} 

Self-supervised pre-training has become a popular approach in natural language processing (NLP) because it allows the creation of high-performance language models. By training a model on a large corpus of text, the model can learn useful representations of the language. However, the effectiveness of these representations is closely related to the type of data used for pre-training. 
When applied to a different type of text, such as a different language, the model may not perform as well, requiring the development of specialized models. Consequently, language-specific models, such as those for German \cite{scheible2020gottbert}, French \cite{martin2019camembert}, or Spanish \cite{canete2020spanish}, have been developed and have been successful in improving performance for these languages. However, even within a language, technical languages can be very different from spoken language, necessitating the development of domain-specific models \cite{gu2021domain}. Medical terminology is a prime example. 

Medical language models that are specifically trained to process and structure medical text have the potential to greatly improve the efficiency and accuracy of medical document analysis. 
However, the challenge in training such models lies in the limited availability of relevant text data, especially for languages other than English. In addition, the sensitive nature of medical information often limits the generation of large medical text corpora. 
Despite these challenges, the development of medical language models is highly desirable, as they could help process and structure the vast amounts of text generated daily in hospitals.

For English, specialized models for biomedical language processing have already been developed \cite{gu2021domain,lee2020biobert}. \cite{lee2020biobert} proposed BioBERT, a version of the BERT model trained on English biomedical abstracts, which outperforms previous models on biomedical benchmarks while maintaining the performance of the original BERT model \cite{devlin2019bert}. Med-BERT, proposed by\cite{rasmy2021med}, is the first model to be fully trained using hospital data, specifically semi-structured EHRs, resulting in improved performance for downstream prediction models. For non-English languages, the development of specialized 
models is more difficult due to less available data. Nevertheless, previous work on German medical models has demonstrated the potential of using specialized models for biomedical language processing \cite{lentzen2022critical}, \cite{frei2022gernermed}, \cite{bressem2020highly}. For example, BioGottBERT, \cite{lentzen2022critical}, which is trained on open medical German texts from Wikipedia and scientific abstracts, could outperform its generalized counterpart \cite{scheible2020gottbert} on medical tasks. However, these models often suffer from limitations such as limited training data, narrow focus (i.e., focused on only one medical subspecialty), or unrepresentative benchmarks, which limit their comparability.

To address these challenges, our goal is to build a comprehensive German clinical language model - \emph{medBERT.de} - that is trained on a diverse set of medical texts, including scientific texts, medical books, and hospital data from various medical domains. Furthermore, by providing openly available benchmarks using realistic hospital data, we aim to enable the reproducibility of our results and facilitate comparison with other models.

\section{Material and Methods} 

This study was approved by the local ethics committees of \emph{Charité - Universitätsmedizin Berlin} (EA2/078/22). In accordance with local laws and regulations, written informed consent was not required due to the retrospective design. A total of 4.7 million documents from 11 different sources were included, representing 10 GB of raw data. For details, see table \ref{table:datasources} or appendix 
\ref{Appendix:Data Sources}. 

\begin{table}
    \centering
    \caption{Data Sources}
    \label{table:datasources}
    \begin{tabular}{lrrrr}
        \toprule
        \textbf{Source}                         & \textbf{No. Documents}    & \textbf{No. Sentences}    & \textbf{No. Words}        & \textbf{Size (MB)}\\
        \midrule
        DocCheck Flexikon                                       & 63,840                    & 720,404                   & 12,299,257                & 92 \\
        \textsc{GGPOnc} 1.0 \cite{borchert-etal-2020-ggponc}    & 4,369                     & 66,256                    & 1,194,345                 & 10 \\
        Webcrawl \cite{Shrestha2021}                            & 11,322                    & 635,806                   & 9,323,774                 & 65 \\
        PubMed abstracts                                        & 12,139                    & 108,936                   & 1,983,752                 & 16 \\
        Radiology reports                                       & 3,657,801                 & 60,839,123                & 520,717,615               & 4,195 \\
        Spinger Nature                                          & 257,999                   & 14,183,396                & 259,284,884               & 1,986 \\
        Electronic health records  \cite{schmidt2021tbase}      & 373,421                   & 4,603,461                 & 69,639,020                & 440 \\
        Doctoral theses                                         & 7,486                     & 4,665,850                 & 90,380,880                & 648 \\
        Thieme Publishing Group                                 & 330,994                   & 10,445,580                & 186,200,935               & 2,898 \\
        Wikipedia                                               & 3,639                     & 161,714                   & 2,799,787                 & 22 \\
        \midrule
        \textbf{Summary}                                        & \textbf{4,723,010}        & \textbf{96,430,526}       & \textbf{1,153,824,249}    &  \textbf{10,372} \\
        \bottomrule
    \end{tabular} \\
\end{table}


\subsection{Data Annotation and Benchmarking}

\subsubsection{Radiology Benchmarks}
Three medical benchmarks, each based on 2000 radiology reports, were created from radiology reports. Patients that undergo radiological examinations often have a variety of different conditions. For this reason, radiology reports capture a greater diversity of information than, e.g., electronic health records that stem from a single medical field. In this study, we focus on three tasks. The first task is a classification task based on short text reports from chest X-rays. The second task, also a classification task, is based on longer reports of CT examinations. The third task is a named entity recognition task (NER), that is based on medium-sized reports of CT/X-ray examinations of the wrist. All of these reports were obtained from a large, level 1 hospital in Germany and cover a wide range of bone, lung, heart, and vascular diseases. 

For the first medical benchmark, a multi-label text classification task, three board-certified radiologists (KKB, LCA, SMN) manually labeled 2000 reports for the global presence/absence of four pathologies and four different types of therapy devices. The second benchmark, a text classification task, contained 2000 reports of computed tomography (CT) scans. All reports had to include the chest but could include additional body parts (head, neck, abdomen).  All studies were labeled by a final-year medical student (LX) for the presence/absence of 23 different chest pathologies and were then reviewed by two board-certified radiologists (LCA, KKB). 
For the third benchmark, a named entity recognition task, 2000 reports were labeled by a final-year medical student (JPL) for the presence/absence of 42 labels. The labels were then verified by two board-certified radiologists (LCA, KKB). Details of each label are provided in the appendix. \ref{appendix:labels}
We make these benchmarks openly available at \url{https://doi.org/10.5281/zenodo.7574287}

\subsubsection{Private Medical Benchmarks}
A significant proportion of radiology reports in the training corpus can lead to overfitting, hindering the model's ability to generalize to new clinical scenarios. Therefore, the benchmarks based on radiological texts alone may not be sufficient to assess how well the models would perform on new types of clinical data. To mitigate this, we developed benchmarks using new, unseen clinical records, namely surgical reports and discharge summaries. Both types of clinical records were not previously included in the training data. The benchmarks were multi-label classification tasks, where the model had to predict multiple diagnoses (ICD-10) or procedure (OPS) codes from the text, allowing a thorough evaluation of its performance in real-world situations. 
Discharge notes and operative reports are more challenging benchmarks than radiology reports because they tend to be longer and more complex. In addition, the information contained in discharge summaries and operative reports differs from that found in radiology reports. While radiology reports focus primarily on describing anatomy, pathological findings, and formulating a diagnosis, discharge notes and operative reports provide a broader context and include information about the patient's condition, treatment, medications, and follow-up plans.

Discharge summaries, surgery reports, as well as all ICD-10 and OPS codes were extracted from the hospital information system of the same level I hospital as before. For these tasks, no manual labeling was performed. Instead, we assigned to the surgery reports as labels all OPS codes of the same patient that matched the date of the text document. Furthermore, we restricted the codes to the surgery chapter of the OPS 
 system \footnote{\url{https://www.dimdi.de/static/de/klassifikationen/ops/kode-suche/opshtml2022/chapter-5.htm}}. For the discharge summaries, we assigned all codes (in one task diagnoses as ICD-10, in another task procedures as OPS) of the patient as labels. For each of these tasks, we included as labels the most frequent codes, such that the test set consisted of at least 10 examples for each label. 
Since all three tasks involve automatic and, particularly for the discharge summaries tasks, only approximately accurate labels (e.g., codes are present but the information could be described elsewhere in the respective EHR), performance even for perfect pattern recognition is not expected to be very high - by construction. Thus, the results of these tasks should not be interpreted in an absolute manner but rather relatively by comparing the performance between models. 
The OPS code 5-984 ("Microsurgical technique") is the most common label used for surgery reports, 8-930 ("Monitoring of respiration, heart, and circulation without measurement of pulmonary artery pressure and central venous pressure") for OPS-labeled discharge summaries, and the ICD code Z11 ("Special procedures for testing for infectious and parasitic diseases") for ICD-10-labeled discharge summaries.

\subsubsection{Open Medical Benchmarks}
As an additional openly available benchmark task, we consider the prediction of named entities in \textsc{GGPOnc} 2.0, a corpus of German clinical practice guidelines in oncology and the largest freely distributable data set of semantically annotated German medical texts \cite{borchert2022ggponc}. Seven medical students (all passed their first medical exam) annotated more than 200K mentions of clinically named entities. For the benchmark, we consider the most challenging setting with 8 fine-grained semantic classes and long entity spans. We use the same training/development/test splits as in the original baseline NER experiments. We also consider \textsc{GraSCCo}, a synthetic corpus of 62 clinical case reports, as a benchmark \cite{modersohn2022grascco}. Since the initial release of \textsc{GraSCCo} does not include semantic annotations, a single annotator from the \textsc{GGPOnc} annotation team created named entity annotations using the original \textsc{GGPOnc} annotation scheme and instructions. This resulted in 5.8K entity annotations in the long/fine setting.

\subsection{Data Anonymization}
Radiology reports extracted from our hospital database typically do not contain identifiable information such as name or date of birth. However, in rare cases, this information may have been added by the radiologist. Therefore, we used a named entity recognition model for the German language to identify all patient names in the \cite{akbik2018coling} data set. Identified names were then manually verified by two radiologists (KKB, LCA) and subsequently removed from the text. In addition, dates were removed from the text and replaced with wildcards. For benchmarks, each document was reviewed at least three times by the authors (KKB, LCA, JMP, PG) to ensure that no identifiable information was present. 

\subsection{Architectures / Models}

To evaluate the performance of our proposed model, we compare six different models based on the BERT or RoBERTa architecture. 
Two general German language models, two German medical models, and two versions of our pre-trained BERT model:
\begin{itemize}
	\item{GottBERT, which is the current state-of-the-art RoBERTa-based model for German text \cite{scheible2020gottbert, liu2019roberta} }
	\item{A multilingual BERT-based model \cite{devlin2018pretraining}.}
	\item{BioGottBERT, a version of GottBERT fine-tuned to medical texts and German Wikipedia \cite{lentzen2022critical}}
	\item{German-MedBERT, a version of the German BERT \cite{deepset2019} fine-tuned on a crawl of German medical websites.}
    \item{\emph{medBERT.de}, our model pre-trained on 4.7 Million German medical texts}
    \item{\emph{medBERT.de\textsubscript{dedup}}, a variant of \emph{medBERT.de} trained on a slightly smaller corpus where duplicated radiology reports had been removed}
\end{itemize}

\subsection{Deduplication}
Radiology reports are often written in a semi-structured form with very similar sentences. Because of this repetition, the information content of many documents is lower in terms of semantic concepts than other data sources used.
Language models tend to quickly overfit due to these data-inherent properties.
A common strategy to counteract this behavior is to deduplicate the pre-training data \cite{dedup}. Therefore, we measure the cosine distance between all reports by encoding them as bag of word representations. We only keep documents for which there is no other document with a similarity greater than 0.75. Due to computational restriction, we had to limit this approach to short reports only. Still, using this approach, we reduce the number of radiology reports for pre-training from 4,504,167 to 3,657,801 reports. To evaluate the impact of deduplication, 
we report the performance of the two BERT models trained with and without deduplicated data. 

\subsection{Hyperparameters/Pretraining Details}
We pre-train the model using the Lamb optimizer \cite{you2019lamb}. As usual for BERT-based models, we train the model in two phases, the first with a maximum sequence length of 128 and the second with a sequence length of 512. To pre-train our model from scratch, we use the hyperparameters given by \cite{you2019lamb}. In the first phase, we use a learning rate of $6e^{-3}$ and a batch size of 65,536 with 2,000 warm-up steps and a polynomial decaying learning rate for a total of 7,038 steps. In the second phase, we train with a batch size of 32,768 and a maximum learning rate of $4e^{-e}$, 200 warmup steps and a total of 1,563 steps.  
We remove very rare Unicode characters that appear less than three times from our pre-training data. This allows the tokenizer vocabulary to contain more specific sub-words and removes unnecessary tokens from the vocabulary that have an impact on the memory footprint. In addition, we set the number of occurrences required for a word to be included in the vocabulary to 20 to avoid including patient names in the vocabulary that may have been missed during anonymization. 

\subsection{Experimental Design}
We perform a hyperparameter optimization on all downstream tasks with median pruning on either the area under the receiver operating characteristic curve (AUROC, classification tasks) or token F1 (NER tasks). For each model-task combination, we perform 100 runs and tune the learning rate, the batch size, and the number of warm-up steps. Finally, we evaluate the best performing models for which the hyperparaemters are reported in Appendix \ref{appendix:parameters}.

\subsection{Data Availability}
Training data is not available due to privacy concerns. Weights for \emph{medBERT.de} and \emph{medBERT.de\textsubscript{dedup}} can be accessed at \url{https://huggingface.co/GerMedBERT/medbert-512} and \url{https://huggingface.co/GerMedBERT/medbert-512-no-duplicates} respectively. 

\section{Results}

\begin{table}
	\centering
	\caption{Classification Tasks}
	\label{table:clftasks}
	\begin{tabular}{lccccc} 
		\toprule
		\textbf{Model}                    & \textbf{AUROC} & \textbf{Macro F1} & \textbf{Micro F1} & \textbf{Precision} & \textbf{Recall} \\
		\midrule
		\textbf{Chest CT} \\
            GottBERT & 92.48 & 69.06 & 83.98 & 76.55 & 65.92 \\ 
            BioGottBERT & 92.71 & 69.42 & 83.41 & 80.67 & 65.52 \\ 
            Multilingual BERT & 91.90 & 66.31 & 80.86 & 68.37 & 65.82 \\ 
            German-MedBERT & 92.48 & 66.40 & 81.41 & 72.77 & 62.37 \\ 
            \emph{medBERT.de}  & \textbf{96.69} & \textbf{81.46} & \textbf{89.39} & \textbf{87.88 }& \textbf{78.77 }\\ 
            \emph{medBERT.de}\textsubscript{dedup} & 96.39 & 78.77 & 89.24 & 84.29 & 76.01 \\ 
		\midrule
		\textbf{Chest X-Ray} \\  
            GottBERT & 83.18 & 64.86 & 74.18 & 59.67 & 78.87 \\ 
            BioGottBERT & 83.48 & 64.18 & 74.87 & 59.04 & 78.90 \\ 
            Multilingual BERT & 82.43 & 63.23 & 73.92 & 56.67 & 75.33 \\ 
            German-MedBERT & 83.22 & 63.13 & 75.39 & 55.66 & 78.03 \\ 
            \emph{medBERT.de}  &\textbf{ 84.65} & \textbf{67.06} & \textbf{76.20} & \textbf{60.44} & \textbf{83.08} \\ 
            \emph{medBERT.de}\textsubscript{dedup} & 84.42 & 66.92 & 76.26 & 60.31 & 82.99 \\ 
 		\midrule
		\multicolumn{4}{l}{\textbf{ICD-10 code classification on discharge notes }} \\  
            GottBERT & 77.23 & 18.32 & 51.23 & 38.30 & 14.27 \\ 
            BioGottBERT & 78.01 & 17.96 & 50.56 & 35.97 & 13.95 \\ 
            Multilingual BERT & 76.64 & 19.48 & 51.19 & 38.39 & 15.60 \\ 
            German-MedBERT & 75.44 & \textbf{23.41} & 53.63 & 41.39 & \textbf{18.94 }\\ 
            \emph{medBERT.de} & 80.78 & \textbf{23.41} & \textbf{53.84} & \textbf{41.42} & 18.75 \\ 
            \emph{medBERT.de}\textsubscript{dedup} & \textbf{80.84 }& 21.44 & 52.46 & 40.45 & 17.04 \\ 
   		\midrule
		\multicolumn{4}{l}{\textbf{OPS code classification on discharge notes }} \\  
            GottBERT & 71.37 & 16.46 & 39.54 & 29.63& 13.02 \\ 
            BioGottBERT & 69.90 & 15.97 & 38.60 & 35.06 & 12.23 \\ 
            Multilingual BERT & 70.39 & 15.53 & 39.16 & 27.94 & 12.35 \\ 
            German-MedBERT & 71.79 & 15.90 & 38.22 & 29.30 & 12.76 \\ 
            \emph{medBERT.de} & 76.83 & 20.48 & 44.96 & 33.82 & 16.57 \\ 
            \emph{medBERT.de}\textsubscript{dedup} & \textbf{77.07} & \textbf{21.33 }& \textbf{46.33 }& \textbf{39.72 }& \textbf{16.83 }\\ 
        \midrule
		\multicolumn{4}{l}{\textbf{OPS code classification on surgery reports}} \\  
            GottBERT & 89.17 & 52.58 &  64.25 &  61.42 & 48.08 \\ 
            BioGottBERT & 92.88 & 54.82 & 66.27 & 61.38 & 50.98 \\ 
            Multilingual BERT & 91.28 & 65.42 & 72.92 & 71.74 & 61.78 \\ 
            German-MedBERT & 92.50 & 55.21 & 66.32 & 62.37 & 51.24 \\ 
            \emph{medBERT.de} &\textbf{ 94.36} & \textbf{66.39} &\textbf{ 73.95} & 69.67 & \textbf{64.25}\\ 
            \emph{medBERT.de}\textsubscript{dedup} & 93.44 & 65.28 & 73.87 & \textbf{73.19} & 61.09 \\ 
        \midrule
    		\textbf{GermEval-18} \\
      		GottBERT 									& \textbf{86.07} & \textbf{79.19} & \textbf{81.11} & \textbf{80.10} & \textbf{78.58}  \\
    		BioGottBERT                       			& 82.36 & 76.57 & 78.95 & 77.72 & 75.88  \\
    		Multilingual BERT 							& 83.65 & 76.76 & 78.63 & 77.09 & 76.49  \\
    		German-MedBERT                    			& 84.72 & 77.85 & 79.60 & 77.69 & 78.02  \\
    		\emph{medBERT.de}                  			& 81.45 & 74.84 & 77.65 & 76.77 & 73.97  \\
    		\emph{medBERT.de}\textsubscript{dedup} 		& 82.05 & 74.56 & 77.40 & 76.22 & 73.75  \\
		\bottomrule
	\end{tabular}
\end{table}

\begin{table}
	\centering
	\caption{NER Tasks}
	\label{table:nertasks}
	\begin{tabular}{lcccccccc} 
		\toprule
		\textbf{Model} & \textbf{AUROC} & \textbf{AUROC\textsubscript{tok}} & \textbf{F1\textsubscript{mac}} & \textbf{F1\textsubscript{tok}} & \textbf{Prec}  & \textbf{Prec\textsubscript{tok}} & \textbf{Rec} & \textbf{Rec\textsubscript{tok}}  \\
		\midrule
		\textbf{Wrist NER} \\
		GottBERT                                & 86.24 & 86.31 & 55.22 & 52.94 & 62.26 & 60.45 & 56.18 & 54.17 \\
		BioGottBERT                             & 87.40 & 87.48 & 54.89 & 52.53 & 61.39 & 59.37 & 57.15 & 54.74 \\
		Multilingual BERT                       & 87.29 & 87.24 & 54.63 & 53.12 & 62.95 & 61.53 & 55.09 & 53.43 \\
		German MedBERT                          & \textbf{88.25} & \textbf{88.31} & 53.40 & 51.26 & 61.33 & 59.77 & 54.54 & 52.50 \\
	    \emph{medBERT.de}                       & 86.84 & 86.88 & \textbf{59.11} & \textbf{58.68} & \textbf{65.66} & \textbf{65.39} & \textbf{60.59} & \textbf{60.06} \\
		\emph{medBERT.de}\textsubscript{dedup}  & 87.45 & 87.40 & 58.15 & 57.03 & 62.88 & 61.87 & 60.26 & 59.11 \\
		\midrule
		\textbf{\textsc{GraSCCo}} \\
		GottBERT                  				& 84.83 & 84.81 & \textbf{76.00} & \textbf{75.75} & \textbf{76.72} & \textbf{76.50} & \textbf{75.42} & 75.06 \\
	    BioGottBERT                  			& 84.82 & 84.56  & 75.52 & 74.95 & 76.37 & 75.96 & 74.84 & 74.08\\
		Multilingual BERT                 		& 83.90 & 83.73  & 72.54 & 71.98 & 72.93 & 72.56 & 72.22 & 71.44\\
		German-MedBERT                 			& 84.03 & 83.95  & 73.14 & 72.85 & 73.05 & 72.86 & 73.41& 73.00\\
    	\emph{medBERT.de}                  		& \textbf{85.14} & \textbf{85.07} & 75.60 & 75.46 & 75.91 & 75.78  & 75.33& \textbf{75.17} \\
		\emph{medBERT.de}\textsubscript{dedup} 	& 84.89 & 84.83  & 75.20 & 75.25 & 75.85 & 75.97 & 74.72 & 74.66\\
		\midrule
		\textbf{\textsc{GGPOnc}} \\
		GottBERT                  				& 98.01 & 98.07 & 75.18 & 73.43 & 76.40 & 74.96 & 74.43& 72.42 \\
		BioGottBERT                  			& 97.96 & 98.05 & \textbf{76.07} & 74.62 & 77.06& 75.45 & \textbf{75.60 }& 74.18 \\
		Multilingual BERT                 		& 97.81 & 97.95 & 73.75 & 72.20 & 75.47 & 73.93 & 72.95 & 71.11\\
		German-MedBERT                 			& 97.55 & 97.66 & 74.48 & 72.45 & 75.59 & 73.73 & 73.76 & 71.57 \\
		\emph{medBERT.de}                  		& \textbf{98.20 } & \textbf{98.35} & 75.93 & \textbf{75.12} & \textbf{77.37} & \textbf{76.33} & 75.15 &\textbf{74.45}\\
		\emph{medBERT.de}\textsubscript{dedup}  & 98.10 & 98.22  & 75.57 & 74.92 & 76.74 & 75.98 & 75.05& 74.43\\
		\bottomrule
	\end{tabular}
\end{table}

\subsection{Radiology Benchmarks} 
In the chest x-ray task, we found that the two best performing models were our own pre-trained BERT models. Our model trained on the corpus with duplicates removed (medBERT.de\textsubscript{dedup}) achieves a slightly better performance with an average AUROC of 83.65 compared to 83.42 of the model trained on the whole corpus (medBERT.de). The third-best performance is achieved by GottBERT with a mean AUROC of 83.48, which also outperformed other medical models. 
In the chest CT task, our model trained on the whole corpus performed best with an AUROC of 96.69, closely followed by  \emph{medBERT.de\textsubscript{dedup}.de}  (AUROC 96.39). Other models show a significantly lower performance of 19\%. GottBERT and BioGottBERT both perform similary with mostly sub-percentage differences in all metrics except precision. 
For the NER task, emph{medBERT.de\textsubscript{dedup}.de} showed the best performance in all metrics except global Recall. However, the scores of all models are in a similar range, with mostly 1-3\% differences between the best and worst-performing models. 
The results suggest that the advantage of domain-specific models is more pronounced for longer texts. On the X-ray (98 ± 27 words) and NER (108 ± 41 words) tasks, which consist of short reports of only a few sentences, the difference between the models is not as pronounced as on the CT reports, which are considerably longer (258 ± 100 words). 

Tables \ref{table:clftasks} and \ref{table:nertasks} provide an overview of all tasks and metrics. Detailed metrics for each class on the radiology benchmarks can be found in the Appendix \ref{appendix:detailed_metrics}.

\subsection{Open Medical Benchmarks}
The \textsc{GGPOnc} benchmark \cite{borchert2022ggponc}, a German corpus based on clinical practice guidelines for oncology, and the \textsc{GraSCCo} benchmark \cite{grassco2022}, a corpus consisting of artificially generated electronic health records for various diseases, were used for this comparison.
On \textsc{GGPOnc}, our models achieved higher AUROC, precision (global and token-level) and token-level recall than the other models, while BioGottBERT achieved the highest macro F1 score and recall. Deduplication of training data did not seem to have a positive impact on model performance, as \emph{medBERT.de} consistently achieved higher metrics than \emph{medBERT.de}\textsubscript{dedup}.
On \textsc{GraSCCo}, GottBERT and \emph{medBERT.de} showed the best performance. \emph{medBERT.de} had the highest AUROC (85.14), AUROC\textsubscript{tok} (75.17), and Recall\textsubscript{tok} (75.17), while GottBERT had the best performance on all other metrics. However, the overall difference between the models is small. 

\subsection{Private Medical Benchmarks}
Three medical benchmarks were constructed using discharge letters and surgical reports, which are not publicly available for privacy reasons. Each benchmark consisted of 2000 texts, which were stratified into a training set of 1000 texts, a validation set of 500 texts, and a test set of 500 texts.
The first task is to predict 65 different ICD-10 codes from discharge summaries. Both of our models show superior performance on this task, outperforming all other models. The best model is \emph{medBERT.de}, although the difference to \emph{medBERT.de\textsubscript{dedup}} is in the sub-percentage range. 
The second task is to predict 49 OPS codes from discharge summaries. Again, \emph{medBERT.de} and \emph{medBERT.de\textsubscript{dedup}.de}  performs better than all other models. However, the overall performance of all models is below average, with particularly low scores for the F1 measure and recall.
In the third task, the classification of 10 OPS codes from surgical reports, our domain-specific models again show the best performance. Compared to the other two tasks, the overall scores are also higher. We attribute this to the fact that this task is less complex since surgical reports are generally shorter and have less variability than discharge summaries. In addition, the number of labels is reduced, which further contributed to the reduced difficulty.

\subsection{General Domain Benchmarks}
In addition to the medical and radiological benchmark tasks, we also evaluated performance on the GermEval18 \cite{germeval18}. We found that GottBERT outperformed all domain-adapted models in all evaluated metrics.
We also observed that the \emph{medBERT.de} models were outperformed by the BioGottBERT model. This can be attributed to the fact that the pre-training corpus for GottBERT contains a greater amount of general domain knowledge and language, giving the model an advantage in general domain tasks. In addition, BioGottBERT, which is based on GottBERT, may have the same advantage but may be affected by catastrophic forgetting because it has only been trained on corpora from the medical domain.

\subsection{Tokenizer Fertility}
Ruse et al. suggests that a tokenizer that produces fewer sub-words per word may improve performance due to better-developed embeddings \cite{rust2019howgoodtokenizer}. Therefore, we measure tokenizer fertility, which measures the average number of sub-word per tokenized word for all of our evaluated models. For the evaluation, we use the text data from our chest CT, chest x-ray classification, and wrist NER tasks. 
As expected, we measure the lowest fertility for the tokenizer of the \emph{medBERT.de} model, which is trained on data following a similar distribution.
We find that tokenizer fertility does not necessarily correlate with improved performance. For example, our \emph{medBERT.de} models both have the lowest fertility score of 1.18 (see table \ref{tab:fertility}) and are the best-performing models on the chest CT and x-ray as well as the wrist NER task. However, it is closely followed by GottBERT, which has a much higher fertility score of 1.75, but performs similarly well. Furthermore, we observe that deduplication of the training data has almost no effect on the fertility of the \emph{medBERT.de} model.

\begin{table}
\centering
\caption{Overview of the tokenizer fertility for the evaluated models measured on the chest-ct, chest X-ray classfication and wrist NER tasks.}
\label{tab:fertility}
\begin{tabular}{ll} 
\toprule
\textbf{Model} & \textbf{Fertility} \\
\midrule
GottBERT & 1.75 \\
BioGottBERT & 1.75 \\
Multilingual BERT & 1.97 \\
German-MedBERT & 1.86 \\
medBERT.de & 1.18 \\
medBERT.de\textsubscript{dedup} & 1.18 \\
\bottomrule
\end{tabular}

\end{table}

\section{Discussion}
In this study, we trained a domain-specific German BERT model on a large data set of German medical texts, including articles, papers, and electronic medical records. We then fine-tuned the model on various medical benchmarks and found that it outperformed both general domain language models and other medical domain models, demonstrating the model's ability to capture the unique characteristics and terminology of German medical language more effectively than general German models. Our results highlight the advantages of using domain-specific language models for the German medical language but also the importance of a large training corpus. In line with previous research \cite{DBLP:conf/emnlp/Perez-MayosBW21}, we attribute the superior performance of our model on medical benchmarks to the larger amount of data used in training compared to German-MedBERT or BioGottBERT. However, our results also suggest that data for pre-training or fine-tuning domain-specific models should not consist solely of specialized language, as this may negatively affect the model's performance on general tasks. This is demonstrated by the Germeval18 task, where GottBERT outperformed BioGottBERT, even though the two models have identical architectures and BioGottBERT was initialized with GottBERT's weights. This performance difference can be attributed to the fact that BioGottBERT was trained on the entire German Wikipedia, which contains general domain language. 
In addition, we observe a drastically improved performance on the OPS and ICD code classification tasks compared to all other domain-specific and general domain models. This indicates that our models are able to generalize to different subdomains that are not included in the pre-training data. The results, therefore, suggest, that the models are able to capture relevant clinical and medical concepts, thus demonstrating their ability to adapt to the different clinical language used in discharge notes.
In contrast to \cite{rust2019howgoodtokenizer}, we did not observe a direct correlation between the fertility of the tokenizer and the performance of the model on downstream tasks. This suggests that fertility alone is not a predictive measure of a model's performance on specialized downstream tasks. Nevertheless, it is likely that the fertility of the tokenizer played a role in the model's performance on tasks involving longer texts, particularly on clinical benchmarks based on discharge notes and surgical reports. Since the texts for these benchmarks were truncated to fit into 512 tokens, some information may have been lost in the process. A more efficient tokenizer may be able to encode more information, potentially improving the model's performance on these tasks.
In our study, we found a mixed impact of data deduplication. While earlier research suggested benefits from deduplication \cite{dedup}, we did not see a consistent improvement in performance with our model (\emph{medBERT.de}) compared to the deduplicated version (\emph{medBERT.de\textsubscript{dedup}}). While on certain benchmarks, our \emph{medBERT.de} performed better than the deduplicated version, on others it performed worse. This discrepancy could be due to the fact that our deduplication process was not as extensive, as it was only applied to short reports. Furthermore, we have not yet applied deduplication to non-radiological text, which might contain duplicates.

\subsection{Limitations} 
A limitation of our study is that about 40\% of the data consists of radiology reports, which may differ in style from other types of electronic health records. In addition, certain medical specialties, such as ophthalmology and pathology, are underrepresented in our sample due to their limited use of radiological imaging. On the other hand, other specialties, such as psychiatry, may be underrepresented because the conditions they treat are not typically seen on imaging. It is also worth noting that the texts were collected from a single university hospital, and it is possible that the performance of our model on new data may be affected by differences in reporting styles between institutions. We suspect that our training data does not sufficiently capture the semantic information needed to improve performance compared to a language model trained on a general domain corpus. This can be partly explained by the repetitive nature of radiology reports.

\subsection{Conclusion and Outlook} 
In conclusion, this study has shown the benefits of using a domain-specific German BERT model, trained on a large data set of German medical texts, for tasks related to the German medical language. The model achieved superior performance compared to the general domain and other medical domain models, underlining the value of using domain-specific models. However, to further improve performance, a future German clinical language model should be trained on on a more diverse data set, e.g., including discharge summaries from a broad range of medical specialties. Nevertheless, this model represents a new state-of-the-art for German clinical language, outperforming GottBERT.

\section{Acknowledgements}
We would like to thank DocCheck AG and Thieme Medical Publishers for their assistance with data collection. We would also like to thank Manjil Shrestha and Rolf Becker for providing additional training data from their German MedBERT model. We also thank the Scientific Computing of the IT Division at the Charité - Universitätsmedizin Berlin for providing computational resources that contributed to the research results reported in this paper.
KKB is grateful for his participation in the BIH Charité Digital Clinician Scientist Program funded by Charité-Universitätsmedizin Berlin and the Berlin Institute of Health. 

\bibliographystyle{abbrv}

\bibliography{bibliography}

\section{Appendix}
\subsection{Data Sources Details} \label{Appendix:Data Sources}

\paragraph{DocCheck Flexikon:} 
The DocCheck Flexikon (\url{https://flexikon.doccheck.com/}) is an open wiki dealing with medical topics. It contains overview articles about diseases, diagnostic procedures, or treatments in all areas of medicine. This study includes all articles of the Flexikon that have been published until January 1\textsuperscript{st}, 2022. In addition, entries from the \textit{DocCheck} forum and product descriptions from the medical store were included. This resulted in 63,884 documents (92 MB of raw text).  

\paragraph{\textsc{GGPOnc}:} 
\textsc{GGPOnc} is a freely available German language corpus based on clinical practice guidelines for oncology with expert annotations. It is available at \cite{borchert-etal-2020-ggponc, borchert2022ggponc}

\paragraph{Webcrawl:} 
A webcrawl of several German medical forums was performed, as described in \cite{Shrestha2021}. The webcrawl consisted of 11,322 documents (65 MB of raw text). 

\paragraph{Pubmed Abstracts:} 
We crawled PubMed (\url{https://pubmed.ncbi.nlm.nih.gov/}) for German publications with openly available German abstracts published by September 1\textsuperscript{st} 2022. This resulted in 12,139 documents, representing 16 MB of raw text.

\paragraph{Radiology Reports:} 
All radiology reports created between January 1\textsuperscript{st}, 2009, and December 31\textsuperscript{st}, 2021, were extracted from the Radiology Information System of \textit{Charité - Universitätsmedizin Berlin}. After removing texts with less than 100 characters, we performed a similarity analysis to exclude duplicate reports. All remaining documents were then anonymized. This resulted in 3.66 million radiology reports that were included in the training corpus (4,195 MB of raw text). 

\paragraph{Springer Nature Corpus:} 
Using the \textit{Springer Nature API}, we identified German-language, open-access publications from the Springer Nature group. Abstracts and full text of the publications were extracted and added to the training corpus. In total, this involved 257,999 documents and 1,986 MB of data. 

\paragraph{Thieme Publishing Group Corpus:} 
With the permission of the \textit{Thieme Publishing Group}, medical textbooks, licensed by the Charité, and journals for continuing medical education (including the Up2Date series, available at \url{https://www.thieme.de/de/aerzte-in-weiterbildung/up2date-fachzeitschriften-20280.htm}) were included in the training corpus. After cleaning the texts by removing figures and tables, this resulted in 330,994 documents and 2,898 MB of raw text. 

\paragraph{Electronic Health Records:} 
We included 373,421 electronic health records (EHR) from the TBase database form the Department of Nephrology and the Center for Kidney Transplantation at \textit{Charité - Universitätsmedizin Berlin} \cite{schmidt2021tbase}. These included physician letters, microbiology reports, pathology reports, and diagnostic procedure reports (440 MB of raw text). 

\paragraph{PhD Theses:} 
In this study, we used a data set of 7,481 open-access German medical dissertations and postdoctoral theses from the \textit{Charité - Universitätsmedizin Berlin} available at \url{https://refubium.fu-berlin.de/handle/fub188/13}. We cleaned the data by removing sentences that did not contain German stop words and excluded theses with a length of fewer than 15 pages. This process ensured that only relevant, high-quality information was included in our analysis. In total, 648 MB of data was added to the training corpus. 

\paragraph{Wikipedia:} 
Entries from the German Wikipedia dealing with medical topics were extracted and added to the training corpus. There were 3,639 texts, corresponding to 22 MB. 

\subsection{Detailed metrics per class for benchmarks based on radiology reports} \label{appendix:detailed_metrics}

In evaluating the performance of our models on radiology report benchmarks, we analyzed detailed metrics per class to provide a deeper understanding of the model's performance on specific classes within the data set. The metrics included precision, recall, F1 score, and AUROC, which were calculated for each class individually, allowing for a more nuanced evaluation of the model's performance and potential shortcomings. 

\begin{longtable}{l|cccc|cccc}
    \caption{Per class metrics for X-ray report classification task} \\
        \hline
        \textbf{Class} & \textbf{AUROC} & \textbf{F1} & \textbf{Precision} & \textbf{Recall} & \textbf{AUROC} & \textbf{F1} & \textbf{Precision} & \textbf{Recall}\\
        \hline
        \endhead
        & \multicolumn{3}{c}{\textbf{GottBERT}} & & \multicolumn{4}{c}{\textbf{BioGottBERT}}  \\
            Congestion & 65.85 & 36.23 & 27.12 & 54.55 & 66.49 & 35.63 & 27.67 & 50.00 \\ 
            Effusion & 76.30 & 60.20 & 50.43 & 74.68 & 77.11 & 61.27 & 51.05 & 76.58 \\ 
            Opacitiy & 66.20 & 45.36 & 36.07 & 61.11 & 68.75 & 47.60 & 36.40 & 68.75 \\ 
            Pneumothorax & 84.85 & 25.12 & 14.53 & 92.86 & 84.01 & 24.07 & 13.83 & 92.86 \\ 
            Gastric tube & 89.92 & 76.57 & 70.53 & 83.75 & 91.06 & 81.10 & 72.20 & 92.50 \\ 
            Thoracic drain & 94.00 & 85.87 & 82.56 & 89.44 & 93.46 & 84.46 & 79.13 & 90.56 \\ 
            Tracheal tube & 89.19 & 84.58 & 82.63 & 86.64 & 87.56 & 84.93 & 82.20 & 87.85 \\ 
            Venous catheter & 92.79 & 91.57 & 91.30 & 91.84 & 91.59 & 91.91 & 91.12 & 92.71 \\ 
            Misplaced & 89.52 & 78.26 & 81.82 & 75.00 & 91.26 & 66.67 & 77.78 & 58.33 \\  
        \hline
        & \multicolumn{3}{c}{\textbf{Multilingual BERT}} & & \multicolumn{4}{c}{\textbf{German-MedBERT} }  \\
            Congestion & 66.31 & 37.61 & 30.14 & 50.00 & 66.18 & 36.63 & 27.03 & 56.82 \\ 
            Effusion & 77.96 & 61.58 & 52.70 & 74.05 & 76.11 & 61.73 & 51.71 & 76.58 \\ 
            Opacitiy & 64.80 & 41.90 & 35.05 & 52.08 & 67.57 & 49.23 & 39.02 & 66.67 \\ 
            Pneumothorax & 82.14 & 23.45 & 14.53 & 60.71 & 82.65 & 22.89 & 13.77 & 67.86 \\ 
            Gastric tube & 88.31 & 79.89 & 71.43 & 90.62 & 91.39 & 80.65 & 71.50 & 92.50 \\ 
            Thoracic drain & 90.93 & 84.78 & 82.98 & 86.67 & 93.05 & 85.57 & 79.81 & 92.22 \\ 
            Tracheal tube & 88.54 & 84.89 & 80.43 & 89.88 & 87.75 & 85.60 & 82.40 & 89.07 \\ 
            Venous catheter & 91.16 & 90.41 & 90.14 & 90.67 & 92.07 & 92.53 & 91.22 & 93.88 \\ 
            Misplaced & 91.71 & 64.52 & 52.63 & 83.33 & 92.21 & 53.33 & 44.44 & 66.67 \\ 
        \hline
        & \multicolumn{3}{c}{\textbf{\textit{medBERT.de}}} & & \multicolumn{4}{c}{\textbf{\textit{medBERT.de\textsubscript{dedup}}}}  \\
            Congestion & 65.02 & 37.04 & 27.47 & 56.82 & 68.90 & 38.55 & 28.34 & 60.23 \\ 
            Effusion & 78.65 & 61.58 & 51.49 & 76.58 & 78.54 & 61.78 & 52.68 & 74.68 \\ 
            Opacitiy & 70.88 & 49.25 & 38.58 & 68.06 & 67.00 & 48.47 & 38.31 & 65.97 \\ 
            Pneumothorax & 86.33 & 24.64 & 14.21 & 92.86 & 84.90 & 25.84 & 14.92 & 96.43 \\ 
            Gastric tube & 90.88 & 80.55 & 71.71 & 91.87 & 91.07 & 80.11 & 72.59 & 89.38 \\ 
            Thoracic drain & 94.49 & 88.08 & 82.52 & 94.44 & 94.10 & 88.31 & 82.93 & 94.44 \\ 
            Tracheal tube & 90.19 & 86.55 & 83.46 & 89.88 & 90.30 & 86.33 & 83.40 & 89.47 \\ 
            Venous catheter & 93.45 & 92.53 & 91.22 & 93.88 & 93.05 & 92.87 & 92.73 & 93.00 \\ 
            Misplaced & 91.93 & 83.33 & 83.33 & 83.33 & 91.93 & 80.00 & 76.92 & 83.33 \\  
        \hline
\end{longtable}
\begin{longtable}{l|cccc|cccc}
    \caption{Per class metrics for CT report classification task} \\
        \hline
        \textbf{Class} & \textbf{AUROC} & \textbf{F1} & \textbf{Precision} & \textbf{Recall} & \textbf{AUROC} & \textbf{F1} & \textbf{Precision} & \textbf{Recall}\\
		\hline
        \endhead  
        & \multicolumn{3}{c}{\textbf{GottBERT}} & & \multicolumn{4}{c}{\textbf{BioGottBERT}}  \\
            Aortic dissection & 88.71 & 0 & 0 & 0 & 98.99 & 0 & 0 & 0 \\ 
            Bone tumors & 75.26 & 0 & 0 & 0 & 76.80 & 14.81 & 40.00 & 9.09 \\ 
            Bronchial abnorm. & 93.18 & 82.24 & 86.27 & 78.57 & 98.04 & 81.36 & 77.42 & 85.71 \\ 
            Congestion & 93.55 & 76.92 & 88.24 & 68.18 & 87.18 & 41.38 & 85.71 & 27.27 \\ 
            Effusion & 99.10 & 97.16 & 96.86 & 97.47 & 99.28 & 97.79 & 97.48 & 98.10 \\ 
            Emphysema \\
            \ \ \ \ Lung & 96.67 & 92.68 & 92.68 & 92.68 & 95.42 & 92.02 & 92.59 & 91.46 \\ 
            \ \ \ \ Soft tissue & 84.08 & 57.89 & 64.71 & 52.38 & 89.48 & 45.71 & 57.14 & 38.10 \\ 
            Enlarged Heart & 96.65 & 91.30 & 91.30 & 91.30 & 93.50 & 80.99 & 94.23 & 71.01 \\ 
            Fibrosis & 93.53 & 86.75 & 94.74 & 80.00 & 92.03 & 82.35 & 87.50 & 77.78 \\ 
            Fractures & 92.38 & 73.77 & 80.36 & 68.18 & 93.08 & 77.17 & 80.33 & 74.24 \\ 
            Hiatal hernia & 99.67 & 77.78 & 87.50 & 70.00 & 98.44 & 80.00 & 80.00 & 80.00 \\ 
            Lung tumor & 94.69 & 88.37 & 90.05 & 86.76 & 95.12 & 88.04 & 87.05 & 89.04 \\ 
            Lymphadenopathy \\
            \ \ \ \ All & 95.69 & 92.24 & 89.54 & 95.11 & 93.03 & 90.64 & 86.94 & 94.67 \\ 
            \ \ \ \ Malignant & 88.88 & 68.12 & 83.93 & 57.32 & 87.23 & 66.67 & 84.91 & 54.88 \\ 
            Mediastinal tumor & 78.50 & 0 & 0 & 0 & 81.03 & 27.59 & 57.14 & 18.18 \\ 
            Pericard. effusion & 97.96 & 88.89 & 90.32 & 87.50 & 97.41 & 93.55 & 96.67 & 90.62 \\ 
            Pleural abnorm. & 72.27 & 25.00 & 100.00 & 14.29 & 79.55 & 44.44 & 100.00 & 28.57 \\ 
            Pneumonia & 94.10 & 84.25 & 81.56 & 87.12 & 95.51 & 86.14 & 85.19 & 87.12 \\ 
            Pneumothorax & 97.23 & 87.50 & 93.33 & 82.35 & 97.03 & 81.36 & 96.00 & 70.59 \\ 
            Pulm. embolism & 99.48 & 90.62 & 90.62 & 90.62 & 99.48 & 84.06 & 78.38 & 90.62 \\ 
            Therapy devices & 97.47 & 92.04 & 95.68 & 88.67 & 95.33 & 90.72 & 93.62 & 88.00 \\ 
            Ventilation & 96.50 & 94.02 & 94.02 & 94.02 & 96.21 & 94.05 & 93.55 & 94.57 \\ 
            Others & 93.02 & 29.63 & 57.14 & 20.00 & 90.03 & 32.00 & 80.00 & 20.00 \\ 

        \hline
        & \multicolumn{3}{c}{\textbf{Multilingual BERT}} & & \multicolumn{4}{c}{\textbf{German-MedBERT} }  \\
        
            Aortic dissection & 99.40 & 0 & 0 & 0 & 97.08 & 0 & 0 & 0 \\ 
            Bone tumors & 67.18 & 0 & 0 & 0 & 73.90 & 0 & 0 & 0 \\ 
            Bronchial abnorm. & 94.26 & 80.39 & 89.13 & 73.21 & 95.52 & 80.00 & 81.48 & 78.57 \\ 
            Congestion & 87.23 & 45.71 & 61.54 & 36.36 & 96.58 & 66.67 & 85.71 & 54.55 \\ 
            Effusion & 98.63 & 94.86 & 90.75 & 99.37 & 98.97 & 96.53 & 96.23 & 96.84 \\ 
            Emphysema \\
            \ \ \ \ Lung & 95.74 & 89.16 & 88.10 & 90.24 & 96.12 & 92.59 & 93.75 & 91.46 \\ 
            \ \ \ \ Soft tissue & 84.70 & 57.78 & 54.17 & 61.90 & 86.48 & 52.94 & 69.23 & 42.86 \\ 
            Enlarged Heart & 97.44 & 86.11 & 82.67 & 89.86 & 91.71 & 70.09 & 85.42 & 59.42 \\ 
            Fibrosis & 96.44 & 77.42 & 75.00 & 80.00 & 96.36 & 82.50 & 94.29 & 73.33 \\ 
            Fractures & 90.30 & 70.15 & 69.12 & 71.21 & 91.48 & 72.57 & 87.23 & 62.12 \\ 
            Hiatal hernia & 98.87 & 70.00 & 70.00 & 70.00 & 99.92 & 90.00 & 90.00 & 90.00 \\ 
            Lung tumor & 94.13 & 86.88 & 82.11 & 92.24 & 93.67 & 86.71 & 88.57 & 84.93 \\ 
            Lymphadenopathy \\
            \ \ \ \ All & 93.72 & 90.83 & 87.30 & 94.67 & 95.10 & 91.70 & 90.13 & 93.33 \\ 
            \ \ \ \ Malignant & 88.40 & 67.11 & 74.63 & 60.98 & 88.40 & 59.50 & 92.31 & 43.90 \\ 
            Mediastinal tumor & 73.64 & 25.00 & 40.00 & 18.18 & 76.49 & 0 & 0 & 0 \\ 
            Pericard. effusion & 97.34 & 86.15 & 84.85 & 87.50 & 93.64 & 85.71 & 100.00 & 75.00 \\ 
            Pleural abnorm. & 80.87 & 16.67 & 20.00 & 14.29 & 70.51 & 0 & 0 & 0 \\ 
            Pneumonia & 92.44 & 77.52 & 68.00 & 90.15 & 95.18 & 83.97 & 84.62 & 83.33 \\ 
            Pneumothorax & 98.92 & 88.89 & 96.55 & 82.35 & 97.31 & 85.71 & 93.10 & 79.41 \\ 
            Pulm. embolism & 99.22 & 77.61 & 74.29 & 81.25 & 99.01 & 84.06 & 78.38 & 90.62 \\ 
            Therapy devices & 95.74 & 87.37 & 89.51 & 85.33 & 95.69 & 86.52 & 92.42 & 81.33 \\ 
            Ventilation & 93.84 & 89.30 & 87.89 & 90.76 & 96.93 & 94.28 & 94.54 & 94.02 \\ 
            Others & 88.13 & 51.61 & 72.73 & 40.00 & 95.06 & 51.61 & 72.73 & 40.00 \\ 
            
		\hline
        & \multicolumn{3}{c}{\textbf{\textit{medBERT.de}}} & & \multicolumn{4}{c}{\textbf{\textit{medBERT.de\textsubscript{dedup}}}}  \\
            Aortic dissection & 80.65 & 66.67 & 100.00 & 50.00 & 67.44 & 0 & 0 & 0\\
            Bone tumors & 94.10 & 64.86 & 80.00 & 54.55 & 97.19 & 64.86 & 80.00 & 54.55\\
            Bronchial abnorm. & 96.70 & 91.74 & 94.34 & 89.29 & 98.16 & 85.98 & 90.20 & 82.14\\
            Congestion & 98.60 & 72.22 & 92.86 & 59.09 & 95.02 & 72.22 & 92.86 & 59.09\\
            Effusion & 99.62 & 97.50 & 96.30 & 98.73 & 99.17 & 97.14 & 97.45 & 96.84\\
            Emphysema & & & & & & & & \\
            \ \ \ \ \ \ Lung  & 97.56 & 95.06 & 96.25 & 93.90 & 97.96 & 93.83 & 95.00 & 92.68\\
            \ \ \ \ \ \ Soft tissue & 91.66 & 54.55 & 75.00 & 42.86 & 92.12 & 63.16 & 70.59 & 57.14\\
            Enlarged heart & 99.44 & 93.53 & 92.86 & 94.20 & 99.86 & 95.04 & 93.06 & 97.10\\
            Fibrosis & 96.33 & 82.98 & 79.59 & 86.67 & 97.94 & 87.06 & 92.50 & 82.22\\
            Fractures & 97.71 & 89.55 & 88.24 & 90.91 & 98.09 & 92.31 & 93.75 & 90.91\\
            Hiatal hernia & 99.69 & 73.68 & 77.78 & 70.00 & 99.73 & 73.68 & 77.78 & 70.00\\
            Lung tumor & 97.68 & 93.24 & 92.00 & 94.52 & 98.42 & 93.75 & 91.70 & 95.89\\
            Lymphadenopathy & & & & & & & & \\
            \ \ \ \ \ \ All & 97.06 & 95.03 & 92.44 & 97.78 & 97.01 & 94.02 & 90.53 & 97.78\\
            \ \ \ \ \ \ Malignant & 97.68 & 85.00 & 87.18 & 82.93 & 97.89 & 84.21 & 91.43 & 78.05\\
            Mediastinal tumor & 86.78 & 16.00 & 66.67 & 9.09 & 93.40 & 8.33 & 50.00 & 4.55\\
            Pericard. effusion & 97.96 & 90.00 & 96.43 & 84.38 & 94.20 & 88.14 & 96.30 & 81.25\\
            Pleural abnorm. & 97.36 & 57.14 & 57.14 & 57.14 & 97.12 & 69.23 & 75.00 & 64.29\\
            Pneumonia & 97.43 & 87.86 & 83.11 & 93.18 & 97.46 & 87.54 & 82.55 & 93.18\\
            Pneumothorax & 99.79 & 92.31 & 96.77 & 88.24 & 99.85 & 92.31 & 96.77 & 88.24\\
            Pulm. embolism & 99.83 & 90.62 & 90.62 & 90.62 & 99.94 & 92.06 & 93.55 & 90.62\\
            Therapy devices & 99.00 & 96.00 & 96.00 & 96.00 & 98.86 & 95.39 & 94.16 & 96.67\\
            Ventilation & 97.96 & 94.43 & 92.23 & 96.74 & 97.49 & 95.16 & 94.15 & 96.20\\       
            Others & 98.28 & 66.67 & 84.62 & 55.00 & 98.34 & 70.59 & 85.71 & 60.00\\                
        \hline
\end{longtable}
\begin{longtable}{l|cccc|cccc}
	\caption{Per class, token level metrics for NER task on CT and X-rays of the upper extremity} \\
	\hline
	\textbf{Class}     & \textbf{AUROC} & \textbf{F1} & \textbf{Precision} & \textbf{Recall} & \textbf{AUROC} & \textbf{F1} & \textbf{Precision} & \textbf{Recall} \\
	\hline
	\endhead
	& \multicolumn{3}{c}{\textbf{GottBERT}} & & \multicolumn{4}{c}{\textbf{BioGottBERT}}  \\
        Amput. & 0.0 & 0.0 & 0.0 & 0.0 & 0.0 & 0.0 & 0.0 & 0.0 \\ 
        Carpal Fx & 96.87 & 82.05 & 92.31 & 73.85 & 98.83 & 83.19 & 97.92 & 72.31 \\ 
        Comb. Forearm Fx & 99.93 & 65.12 & 48.28 & 100.0 & 100.0 & 100.0 & 100.0 & 100.0 \\ 
        DISI & 99.6 & 46.67 & 53.85 & 41.18 & 98.89 & 51.85 & 70.0 & 41.18 \\ 
        Disloc. & 99.97 & 86.49 & 76.19 & 100.0 & 99.98 & 85.71 & 78.95 & 93.75 \\ 
        Dist. Rad. Fx & 98.73 & 40.0 & 42.86 & 37.5 & 99.97 & 55.56 & 50.0 & 62.5 \\ 
        Finger Fx & 82.47 & 0.0 & 0.0 & 0.0 & 91.12 & 0.0 & 0.0 & 0.0 \\ 
        Fiss. & 99.29 & 79.16 & 75.38 & 83.33 & 98.63 & 86.02 & 83.33 & 88.89 \\ 
        Hum. Fx & 99.01 & 81.24 & 84.91 & 77.87 & 98.95 & 85.95 & 88.66 & 83.4 \\ 
        Incorr. Mat. Pos. & 95.98 & 78.49 & 100.0 & 64.6 & 98.89 & 70.21 & 88.0 & 58.41 \\ 
        Joint Involv. & 78.21 & 0.0 & 0.0 & 0.0 & 75.43 & 0.0 & 0.0 & 0.0 \\ 
        Lux. & 0.0 & 0.0 & 0.0 & 0.0 & 0.0 & 0.0 & 0.0 & 0.0 \\ 
        Mat. Break & 0.0 & 0.0 & 0.0 & 0.0 & 0.0 & 0.0 & 0.0 & 0.0 \\ 
        Mat. Intact & 100.0 & 80.0 & 66.67 & 100.0 & 100.0 & 83.33 & 71.43 & 100.0 \\ 
        Midhand Fx & 99.99 & 87.5 & 77.78 & 100.0 & 99.99 & 87.5 & 77.78 & 100.0 \\ 
        No Disloc. & 99.39 & 75.86 & 68.75 & 84.62 & 99.71 & 78.05 & 74.42 & 82.05 \\ 
        No Disloc. & 98.66 & 81.1 & 88.06 & 75.16 & 97.98 & 80.66 & 83.11 & 78.34 \\ 
        No Fx & 98.65 & 89.6 & 89.24 & 89.96 & 99.07 & 85.69 & 82.14 & 89.56 \\ 
        No Joint Involv. & 99.63 & 75.0 & 75.51 & 74.5 & 99.23 & 77.85 & 77.85 & 77.85 \\ 
        No Scapholun. Diss. & 99.21 & 78.81 & 75.86 & 81.99 & 97.27 & 78.86 & 76.23 & 81.68 \\ 
        No Ulna Fx & 100.0 & 73.33 & 100.0 & 57.89 & 99.42 & 59.26 & 100.0 & 42.11 \\ 
        Old Fx & 99.87 & 20.0 & 100.0 & 11.11 & 94.17 & 13.79 & 100.0 & 7.41 \\ 
        PISI & 0.0 & 0.0 & 0.0 & 0.0 & 0.0 & 0.0 & 0.0 & 0.0 \\ 
        Pseudoarth. & 100.0 & 90.0 & 85.71 & 94.74 & 99.99 & 89.47 & 89.47 & 89.47 \\ 
        Rad. Fx & 99.85 & 59.34 & 45.0 & 87.1 & 99.94 & 56.25 & 41.54 & 87.1 \\ 
        STT OA & 0.0 & 0.0 & 0.0 & 0.0 & 0.0 & 0.0 & 0.0 & 0.0 \\ 
        Scap. Fx & 0.0 & 0.0 & 0.0 & 0.0 & 0.0 & 0.0 & 0.0 & 0.0 \\ 
        Scaphoid Fx & 98.85 & 76.01 & 73.29 & 78.93 & 98.01 & 74.61 & 77.14 & 72.24 \\ 
        Scapholun. Diss. & 99.18 & 81.02 & 77.72 & 84.62 & 99.34 & 75.81 & 69.46 & 83.43 \\ 
        Sublux. & 100.0 & 98.18 & 96.43 & 100.0 & 100.0 & 100.0 & 100.0 & 100.0 \\ 
        Thumb OA & 99.68 & 0.0 & 0.0 & 0.0 & 99.94 & 28.0 & 16.28 & 100.0 \\ 
        Ulna Fx & 98.0 & 97.19 & 97.08 & 97.31 & 98.02 & 97.05 & 97.15 & 96.96 \\ 
        Ulna+ & 95.44 & 31.82 & 55.26 & 22.34 & 93.68 & 37.04 & 60.98 & 26.6 \\ 
        Ulna- & 98.1 & 71.26 & 68.0 & 74.84 & 98.65 & 65.79 & 56.56 & 78.62 \\ 
        Wrist OA & 100.0 & 100.0 & 100.0 & 100.0 & 100.0 & 100.0 & 100.0 & 100.0 \\ 

	\hline
	& \multicolumn{3}{c}{\textbf{Multilingual BERT}} & & \multicolumn{4}{c}{\textbf{German-MedBERT} }  \\
 
        Amput. & 0.0 & 0.0 & 0.0 & 0.0 & 0.0 & 0.0 & 0.0 & 0.0 \\ 
        Carpal Fx & 97.52 & 68.52 & 84.09 & 57.81 & 95.52 & 64.22 & 77.78 & 54.69 \\ 
        Comb. Forearm Fx & 96.71 & 0.0 & 0.0 & 0.0 & 98.61 & 60.0 & 100.0 & 42.86 \\ 
        DISI & 96.9 & 12.5 & 100.0 & 6.67 & 99.24 & 38.1 & 25.0 & 80.0 \\ 
        Disloc. & 99.97 & 88.89 & 80.0 & 100.0 & 99.98 & 88.89 & 80.0 & 100.0 \\ 
        Dist. Rad. Fx & 98.4 & 46.15 & 60.0 & 37.5 & 99.35 & 42.86 & 50.0 & 37.5 \\ 
        Finger Fx & 94.58 & 0.0 & 0.0 & 0.0 & 88.48 & 0.0 & 0.0 & 0.0 \\ 
        Fiss. & 97.29 & 81.42 & 79.68 & 83.24 & 99.33 & 74.16 & 64.85 & 86.59 \\ 
        Hum. Fx & 98.84 & 82.05 & 82.54 & 81.57 & 97.36 & 79.68 & 80.97 & 78.43 \\ 
        Incorr. Mat. Pos. & 98.88 & 68.97 & 98.36 & 53.1 & 99.8 & 74.89 & 74.56 & 75.22 \\ 
        Joint Involv. & 77.39 & 0.0 & 0.0 & 0.0 & 72.16 & 0.0 & 0.0 & 0.0 \\ 
        Lux. & 0.0 & 0.0 & 0.0 & 0.0 & 0.0 & 0.0 & 0.0 & 0.0 \\ 
        Mat. Break & 0.0 & 0.0 & 0.0 & 0.0 & 0.0 & 0.0 & 0.0 & 0.0 \\ 
        Mat. Intact & 100.0 & 83.33 & 71.43 & 100.0 & 100.0 & 100.0 & 100.0 & 100.0 \\ 
        Midhand Fx & 99.99 & 87.5 & 77.78 & 100.0 & 99.99 & 87.5 & 77.78 & 100.0 \\ 
        No Disloc. & 99.55 & 64.15 & 50.75 & 87.18 & 97.73 & 70.0 & 68.29 & 71.79 \\ 
        No Disloc. & 89.9 & 78.45 & 88.1 & 70.7 & 97.5 & 77.12 & 75.93 & 78.34 \\ 
        No Fx & 96.91 & 87.04 & 85.77 & 88.35 & 99.14 & 83.27 & 79.06 & 87.95 \\ 
        No Joint Involv. & 99.34 & 74.68 & 71.88 & 77.7 & 96.89 & 78.32 & 81.16 & 75.68 \\ 
        No Scapholun. Diss. & 98.38 & 76.83 & 77.56 & 76.1 & 97.74 & 76.48 & 82.78 & 71.07 \\ 
        No Ulna Fx & 99.38 & 0.0 & 0.0 & 0.0 & 92.92 & 0.0 & 0.0 & 0.0 \\ 
        Old Fx & 100.0 & 100.0 & 100.0 & 100.0 & 99.53 & 20.69 & 100.0 & 11.54 \\ 
        PISI & 0.0 & 0.0 & 0.0 & 0.0 & 0.0 & 0.0 & 0.0 & 0.0 \\ 
        Pseudoarth. & 100.0 & 90.0 & 85.71 & 94.74 & 100.0 & 87.8 & 81.82 & 94.74 \\ 
        Rad. Fx & 98.96 & 74.67 & 65.12 & 87.5 & 99.82 & 62.79 & 50.0 & 84.38 \\ 
        STT OA & 0.0 & 0.0 & 0.0 & 0.0 & 0.0 & 0.0 & 0.0 & 0.0 \\ 
        Scap. Fx & 0.0 & 0.0 & 0.0 & 0.0 & 0.0 & 0.0 & 0.0 & 0.0 \\ 
        Scaphoid Fx & 98.15 & 77.03 & 76.51 & 77.55 & 98.06 & 74.96 & 78.44 & 71.77 \\ 
        Scapholun. Diss. & 98.05 & 73.87 & 75.0 & 72.78 & 98.24 & 73.72 & 87.1 & 63.91 \\ 
        Sublux. & 99.63 & 89.8 & 100.0 & 81.48 & 99.98 & 68.29 & 100.0 & 51.85 \\ 
        Thumb OA & 96.86 & 57.14 & 42.86 & 85.71 & 99.69 & 92.31 & 100.0 & 85.71 \\ 
        Ulna Fx & 97.2 & 96.26 & 95.66 & 96.86 & 97.21 & 96.02 & 95.64 & 96.41 \\ 
        Ulna+ & 94.63 & 31.15 & 63.33 & 20.65 & 90.81 & 24.24 & 40.0 & 17.39 \\ 
        Ulna- & 95.84 & 61.35 & 59.88 & 62.89 & 98.09 & 72.73 & 68.13 & 77.99 \\ 
        Wrist OA & 100.0 & 100.0 & 100.0 & 100.0 & 100.0 & 100.0 & 100.0 & 100.0 \\ 
        
	\hline
	& \multicolumn{3}{c}{\textbf{\textit{medBERT.de}}} & & \multicolumn{4}{c}{\textbf{\textit{medBERT.de\textsubscript{dedup}}}}  \\
 
        Amput. & 0.0 & 0.0 & 0.0 & 0.0 & 0.0 & 0.0 & 0.0 & 0.0 \\ 
        Carpal Fx & 97.1 & 69.23 & 90.0 & 56.25 & 98.2 & 83.05 & 90.74 & 76.56 \\ 
        Comb. Forearm Fx & 100.0 & 100.0 & 100.0 & 100.0 & 95.36 & 0.0 & 0.0 & 0.0 \\ 
        DISI & 93.81 & 44.44 & 50.0 & 40.0 & 97.97 & 63.64 & 100.0 & 46.67 \\ 
        Disloc. & 99.96 & 72.73 & 57.14 & 100.0 & 99.98 & 80.0 & 66.67 & 100.0 \\ 
        Dist. Rad. Fx & 99.98 & 66.67 & 53.85 & 87.5 & 99.99 & 72.73 & 57.14 & 100.0 \\ 
        Finger Fx & 48.2 & 0.0 & 0.0 & 0.0 & 84.76 & 0.0 & 0.0 & 0.0 \\ 
        Fiss. & 99.2 & 81.35 & 75.85 & 87.71 & 99.01 & 85.71 & 82.81 & 88.83 \\ 
        Hum. Fx & 98.97 & 84.35 & 82.16 & 86.67 & 97.97 & 83.64 & 86.25 & 81.18 \\ 
        Incorr. Mat. Pos. & 98.97 & 63.16 & 93.1 & 47.79 & 99.35 & 75.14 & 100.0 & 60.18 \\ 
        Joint Involv. & 70.89 & 0.0 & 0.0 & 0.0 & 73.7 & 0.0 & 0.0 & 0.0 \\ 
        Lux. & 0.0 & 0.0 & 0.0 & 0.0 & 0.0 & 0.0 & 0.0 & 0.0 \\ 
        Mat. Break & 0.0 & 0.0 & 0.0 & 0.0 & 0.0 & 0.0 & 0.0 & 0.0 \\ 
        Mat. Intact & 100.0 & 100.0 & 100.0 & 100.0 & 100.0 & 100.0 & 100.0 & 100.0 \\ 
        Midhand Fx & 100.0 & 87.5 & 77.78 & 100.0 & 99.99 & 87.5 & 77.78 & 100.0 \\ 
        No Disloc. & 92.52 & 80.28 & 89.76 & 72.61 & 93.75 & 78.41 & 81.94 & 75.16 \\ 
        No Disloc. & 99.57 & 65.12 & 59.57 & 71.79 & 99.75 & 69.05 & 64.44 & 74.36 \\ 
        No Fx & 98.42 & 87.14 & 83.46 & 91.16 & 97.9 & 87.35 & 84.72 & 90.16 \\ 
        No Joint Involv. & 98.78 & 72.03 & 68.71 & 75.68 & 99.05 & 77.15 & 68.78 & 87.84 \\ 
        No Scapholun. Diss. & 98.07 & 81.63 & 81.5 & 81.76 & 98.14 & 78.34 & 76.58 & 80.19 \\ 
        No Ulna Fx & 99.87 & 77.42 & 100.0 & 63.16 & 100.0 & 78.79 & 92.86 & 68.42 \\ 
        Old Fx & 99.97 & 70.0 & 100.0 & 53.85 & 100.0 & 96.0 & 100.0 & 92.31 \\ 
        PISI & 0.0 & 0.0 & 0.0 & 0.0 & 0.0 & 0.0 & 0.0 & 0.0 \\ 
        Pseudoarth. & 99.99 & 95.0 & 90.48 & 100.0 & 100.0 & 95.0 & 90.48 & 100.0 \\ 
        Rad. Fx & 98.86 & 63.16 & 54.55 & 75.0 & 96.29 & 72.73 & 70.59 & 75.0 \\ 
        STT OA & 0.0 & 0.0 & 0.0 & 0.0 & 0.0 & 0.0 & 0.0 & 0.0 \\ 
        Scap. Fx & 0.0 & 0.0 & 0.0 & 0.0 & 0.0 & 0.0 & 0.0 & 0.0 \\ 
        Scaphoid Fx & 98.19 & 78.32 & 77.41 & 79.25 & 97.9 & 76.45 & 72.7 & 80.61 \\ 
        Scapholun. Diss. & 98.9 & 67.48 & 69.38 & 65.68 & 98.77 & 78.31 & 79.75 & 76.92 \\ 
        Sublux. & 99.99 & 82.61 & 100.0 & 70.37 & 98.89 & 89.8 & 100.0 & 81.48 \\ 
        Thumb OA & 95.96 & 80.0 & 75.0 & 85.71 & 100.0 & 82.35 & 70.0 & 100.0 \\ 
        Ulna Fx & 97.74 & 96.52 & 96.48 & 96.55 & 97.22 & 96.54 & 96.47 & 96.62 \\ 
        Ulna+ & 95.64 & 26.09 & 65.22 & 16.3 & 92.93 & 36.64 & 61.54 & 26.09 \\ 
        Ulna- & 97.95 & 74.01 & 67.18 & 82.39 & 90.87 & 71.52 & 71.97 & 71.07 \\ 
        Wrist OA & 100.0 & 100.0 & 100.0 & 100.0 & 100.0 & 100.0 & 100.0 & 100.0 \\ 
	          
	\hline
\end{longtable}

\subsection{Distribution of classes for benchmarks based on radiology reports} \label{appendix:labels}

For the benchmarks based on radiology reports, the distribution of classes is an important factor to consider when evaluating the performance of our models. As some diagnoses are rarer than others, the class distribution is highly skewed, especially for the CT and NER tasks. 

\centering

\begin{longtable}[ht]{lrrr} \\
\caption{Class distribution for X-ray classification task} \\
  \hline
    Class & Train & Valid & Test \\ 
  \hline
    Congestion & 273 &  89 &  88 \\ 
    Opacitiy & 584 & 120 & 144 \\ 
    Effusion & 460 & 214 & 158 \\ 
    Pneumothorax &  74 &  19 &  28 \\ 
    Thoracic drain & 190 &  80 & 180 \\ 
    Venous catheter & 579 & 283 & 343 \\ 
    Gastric tube & 244 & 117 & 160 \\ 
    Tracheal tube & 397 & 230 & 247 \\ 
    Misplaced &  49 &  20 &  12 \\ 
   \hline
\end{longtable}

\begin{longtable}{lrrr} \\
\caption{Class distribution for CT classification task} \\
  \hline
    Class & Train & Valid & Test \\ 
  \hline
    Aortic dissection &   1 &   2 &   2 \\ 
    Bone tumors &  25 &  19 &  22 \\ 
    Bronchial abnormalities &  98 &  38 &  56 \\ 
    Congestion &  40 &  19 &  22 \\ 
    Effusion & 251 & 138 & 158 \\ 
    Emphysema \\
    \ \ \ \ Lung & 179 &  93 &  82 \\ 
    \ \ \ \ Soft tissue  &  42 &  24 &  21 \\ 
    Enlarged heart & 128 &  62 &  69 \\ 
    Fibrosis &  91 &  45 &  45 \\ 
    Fractures & 118 &  77 &  66 \\ 
    Hiatal hernia &  26 &  10 &  10 \\ 
    Lung tumor & 427 & 224 & 219 \\ 
    Lymphadenopathy \\
    \ \ \ \ All & 396 & 223 & 225 \\ 
    \ \ \ \ Malignant & 154 &  83 &  82 \\ 
    Mediastinal tumor &  33 &  21 &  22 \\ 
    Pericardial effusion &  69 &  32 &  32 \\ 
    Pleural abnormalities &  26 &  15 &  14 \\ 
    Pneumonia & 268 & 119 & 132 \\ 
    Pneumothorax &  49 &  33 &  34 \\ 
    Pulmonary embolism &  49 &  33 &  32 \\ 
    Therapy devices & 243 & 135 & 150 \\ 
    Ventilation & 306 & 165 & 184 \\ 
    Others &  41 &  17 &  20 \\ 
   \hline
\end{longtable}
\begin{longtable}{lrrr} \\
 \caption{Class distribution for NER task on X-ray/CT of the wrist} \\
 \hline
    Class & Train & Valid & Test \\ 
  \hline
    Amput. & 1.00 &  & 1.00 \\ 
    Carpal Fx & 175.00 & 41.00 & 30.00 \\ 
    Comb. Forearm Fx & 9.00 & 2.00 & 4.00 \\ 
    DISI & 17.00 & 2.00 & 1.00 \\ 
    Disloc. & 444.00 & 73.00 & 75.00 \\ 
    Dist. Rad. Fx & 410.00 & 48.00 & 53.00 \\ 
    Finger Fx & 7.00 & 2.00 & 3.00 \\ 
    Fiss. & 5.00 & 2.00 & 3.00 \\ 
    Hum. Fx & 42.00 & 5.00 & 8.00 \\ 
    Joint Involv. & 504.00 & 70.00 & 75.00 \\ 
    Lux. & 14.00 & 4.00 & 2.00 \\ 
    Mat. Intact & 25.00 & 5.00 & 2.00 \\ 
    Midhand Fx & 117.00 & 16.00 & 26.00 \\ 
    No Disloc. & 613.00 & 113.00 & 88.00 \\ 
    No Fx & 751.00 & 106.00 & 91.00 \\ 
    No Joint Involv. & 42.00 & 5.00 & 13.00 \\ 
    No Scapholun. Diss. & 110.00 & 21.00 & 7.00 \\ 
    Old Fx & 75.00 & 10.00 & 13.00 \\ 
    Pseudoarth. & 9.00 & 2.00 & 2.00 \\ 
    Rad. Fx & 112.00 & 14.00 & 13.00 \\ 
    STT OA & 18.00 & 4.00 & 2.00 \\ 
    Scaphoid Fx & 90.00 & 19.00 & 17.00 \\ 
    Scapholun. Diss. & 35.00 & 3.00 & 5.00 \\ 
    Sublux. & 35.00 &  & 3.00 \\ 
    Thumb OA & 47.00 & 7.00 & 12.00 \\ 
    Ulna Fx & 220.00 & 29.00 & 25.00 \\ 
    Ulna- & 22.00 & 2.00 & 3.00 \\ 
    Wrist OA & 6.00 &  & 2.00 \\ 
   \hline
\end{longtable}

\newpage

\subsection{Model Selection for Radiology tasks} \label{appendix:parameters}

For the benchmarks based on radiology reports, the hyperparameter optimization resulted in the following parameters:

\centering

\begin{longtable}[ht]{lrrr} \\
\caption{Parameters of the best models for CT classification task} \\
  \hline
    Model & batch size & learning rate & warmup steps \\ 
  \hline
    GottBERT & 16 &  3.42e-05 &  23 \\ 
    BioGottBERT & 16 & 6.69e-05 & 78 \\ 
    Multilingual BERT &  8 &  2.96e-05 &  648 \\ 
    German-MedBERT & 16 &  3.96e-05 & 146 \\ 
    \textit{medBERT.de} & 8 & 2.41e-05 & 1 \\ 
    \textit{medBERT.de\textsubscript{dedup}} & 8 & 1.9e-05 & 49 \\ 
   \hline
\end{longtable}
\centering

\begin{longtable}[ht]{lrrr} \\
\caption{Parameters of the best models for Xray classification task} \\
  \hline
    Model & batch size & learning rate & warmup steps \\ 
  \hline
    GottBERT & 8 &  2.04e-05 &  454 \\ 
    BioGottBERT & 16 & 3.81e-05 & 293 \\ 
    Multilingual BERT &  16 &  7.77e-05 &  995 \\ 
    German-MedBERT & 8 &  4.20e-05 & 249 \\ 
    \textit{medBERT.de} & 8 & 1.18e-05 & 140 \\ 
    \textit{medBERT.de\textsubscript{dedup}} & 16 & 3.80e-05 & 2 \\ 
   \hline
\end{longtable}
\centering

\begin{longtable}[ht]{lrrr} \\
\caption{Parameters of the best models for NER task on task on CT and X-rays of
the upper extremity} \\
  \hline
    Model & batch size & learning rate & warmup steps \\ 
  \hline
    GottBERT & 8 &  2.18e-05 &  957 \\ 
    BioGottBERT & 8 & 3.81e-05 & 243 \\ 
    Multilingual BERT &  16 &  6.92e-05 &  882 \\ 
    German-MedBERT & 8 &  3.32e-05 & 232 \\ 
    \textit{medBERT.de} & 8 & 4.79e-05 & 466 \\ 
    \textit{medBERT.de\textsubscript{dedup}} & 8 & 2.22e-05 & 40 \\ 
   \hline
\end{longtable}

\end{document}